\documentclass[10pt,twocolumn,letterpaper]{article}

\usepackage{cvpr}              %

\definecolor{cvprblue}{rgb}{0.21,0.49,0.74}
\usepackage[pagebackref,breaklinks,colorlinks,allcolors=cvprblue]{hyperref}
\usepackage{hyperref}
\usepackage{url}
\usepackage{graphicx}
\usepackage{booktabs}
\usepackage{makecell}      %
\usepackage[table]{xcolor} %
\usepackage{array}
\usepackage{tabularx}%
\usepackage{marvosym}
\usepackage{multirow}
\usepackage{adjustbox}
\usepackage{array}
\usepackage{fontawesome5}
\usepackage[accsupp]{axessibility}
\newcolumntype{L}[1]{>{\raggedright\arraybackslash}p{#1}}
\newcolumntype{C}[1]{>{\centering\arraybackslash}p{#1}}
\newcolumntype{Y}{>{\centering\arraybackslash}X}

\newcommand \blfootnote[1]{
    \begingroup
        \renewcommand
        \thefootnote{}\footnote{#1}
        \addtocounter{footnote}{-1}
        \vspace{-1ex}
    \endgroup
}

\def\confName{CVPR}
\def\confYear{2026}

\title{HiF-VLA: Hindsight, Insight and Foresight through Motion Representation \\ for Vision-Language-Action Models}

\author{Minghui Lin$^{1}$, Pengxiang Ding$^{1,2}$\footnotemark[2]\>\,, Shu Wang$^{1}$, Zifeng Zhuang$^{1,2}$, Yang Liu$^{1}$, Xinyang Tong$^{1}$,\\ Wenxuan Song$^{3}$, Shangke Lyu$^{4}$, Siteng Huang$^{2}$\footnotemark[2]\>\,$^{\text{\Letter}}$, Donglin Wang$^{1, 5}$$^{\text{\Letter}}$,\\
{$^1$Westlake University\ \ $^2$Zhejiang University\ \ $^3$HKUST(GZ)\ \ $^4$Nanjing University}\ \ $^5$Westlake Robotics\\%[-2pt]
{\tt\small linminghui@westlake.edu.cn,\ \ siteng.huang@gmail.com}\\[2pt]
\href{https://hifvla.github.io}{\faGlobe\ https://hifvla.github.io} \quad \quad %
\href{https://github.com/OpenHelix-Team/HiF-VLA}{\faGithub\ https://github.com/OpenHelix-Team/HiF-VLA}
}

\begin{document}

\maketitle

\blfootnote{$^\dagger$ Project leads. $^{\text{\Letter}}$ Corresponding authors.}

\begin{abstract}
Vision-Language-Action (VLA) models have recently enabled robotic manipulation by grounding visual and linguistic cues into actions. However, most VLAs assume the Markov property, relying only on the current observation and thus suffering from temporal myopia that degrades long-horizon coherence. In this work, we view motion as a more compact and informative representation of temporal context and world dynamics, capturing inter-state changes while filtering static pixel-level noise. From this perspective, HiF-VLA 
equips a motion-centric world model for the VLA,
enabling agents to reason about temporal dynamics for future evolution during action generation.
Building on this idea, we propose \textbf{HiF-VLA} (\textbf{Hindsight, Insight, and Foresight} for VLAs), a unified framework that leverages motion for bidirectional temporal reasoning. HiF-VLA encodes past dynamics through hindsight priors, anticipates future motion via foresight reasoning, and integrates both through a hindsight-modulated joint expert to enable a ``\textbf{think-while-acting}'' paradigm for long-horizon manipulation. As a result, HiF-VLA surpasses strong baselines on LIBERO-Long and CALVIN ABC-D benchmarks, while incurring negligible additional inference latency. Furthermore, HiF-VLA achieves substantial improvements in real-world long-horizon manipulation tasks, demonstrating its broad effectiveness in practical robotic settings.
\end{abstract}

\section{Introduction}
\label{sec:intro}

\begin{figure*}[ht]
    \centering
    \includegraphics[width=1\linewidth]{img/compare.pdf}
    \caption{(a-b) Comparison with existing methods: VLAs rely on instantaneous observations~\citep{kim2024openvla,kim2025fine} (a-top), stack multiple past frames~\citep{liu2025towards,tianpredictive} (a-second), or generate pixel-level subgoals~\citep{zhao2025cot,zhang2025up} (a-third), suffering from redundancy, high inference cost, and weak structure. In contrast, HiF-VLA (a-bottom) jointly models Hindsight, Insight, and Foresight, expanding the temporal receptive field bidirectionally for compact, structured, and efficient reasoning. (c) HiF-VLA reduces inference latency and achieves state-of-the-art performance on LIBERO-Long and CALVIN ABC-D, significantly outperforming the baseline in real-world experiments.}
    \label{fig:compare}
    \vspace{-10pt}
\end{figure*}

Vision-Language-Action models (VLAs)~\citep{zitkovich2023rt,kim2024openvla,black2024pi_0,intelligence2025pi_,jiang2025rynnvla001,zhen20243d} have emerged as a promising framework for robotic manipulation, leveraging powerful Vision-Language Models (VLMs)~\citep{team2023gemini,chen2023pali,karamcheti2024prismatic,bai2023qwen,beyer2024paligemma} to map visual and language representations to the action space.
Despite progress, most VLAs~\cite{team2024octo,kim2024openvla,kim2025fine} implicitly assume a Markov property, predicting actions solely from the current observation without explicitly modeling temporal dependencies. This simplification leads to a form of \textbf{temporal myopia}. However, long-horizon manipulation requires reasoning that extends beyond the present, maintaining temporal continuity across visual, language, and motor modalities. Without such temporal reasoning, dependencies between consecutive actions deteriorate, resulting in fragmented trajectories and diminished task-level coherence.

Recent efforts~\citep{zhengtracevla,liu2025towards,team2024octo} have sought to alleviate this temporal myopia by incorporating historical context, most commonly by stacking multiple past observation frames. However, this approach is fundamentally limited. Stacking raw frames is not only computationally prohibitive and increases inference latency, hindering real-time control (see \cref{tab:efficient}), but it also introduces substantial pixel-level redundancy. This deluge of static information often obscures the salient, task-relevant dynamics, making it difficult for the model to distinguish meaningful changes from background noise. 
We argue that a more precise and efficient representation of history is not the raw visual content of the past, but the motion that transpired between states. 
Motion serves as a direct and compact proxy for temporal memory and environment dynamics, faithfully capturing the dynamics of interactions (such as an object being moved or a drawer being closed), while discarding redundant static information. This makes motion an ideal primitive for representing history in a way that is both expressive and computationally efficient.

Building on the principle of motion as a compact representation of history, we argue that robust decision-making requires bidirectional temporal reasoning, connecting the past to the future. 
An intelligent agent must possess \textbf{hindsight}, the ability to interpret recent dynamics that led to its current state, grounding its decisions in verified past outcomes. At the same time, it must exhibit \textbf{foresight}, the capability to anticipate plausible future dynamics, enabling proactive, goal-directed behavior rather than purely reactive responses. Such bidirectional reasoning over past and future dynamics is also central to world models.
We identify motion as the natural bridge unifying these two temporal dimensions. Consequently, we propose \textbf{HiF-VLA} (\textbf{Hindsight, Insight, and Foresight} for VLA Models), a unified framework that leverages motion to structure this reasoning process. 
As detailed in \cref{fig:primary}, HiF-VLA comprises three components:
\textbf{1) Hindsight prior acquisition:} encodes historical frames into structured, low-dimensional motion vectors~\citep{jin2024video} as hindsight, preserving dynamics without redundant pixels.
\textbf{2) Foresight reasoning with insight:} interprets task instructions and current observations to anticipate plausible foresight motions and latent action tokens.
\textbf{3) Hindsight-modulated joint expert:} applies hindsight as a top-down constraint on foresight and action streams, which interact within the expert decoder to generate temporally coherent actions.
As shown in \cref{fig:compare}, compared to methods that rely solely on frame stacking or subgoal prediction, HiF-VLA more effectively bridges \emph{perception}, \emph{dynamics}, and \emph{control} within a unified representational space, 
forming a motion-centric World Action Model~\cite{ye2026worldactionmodelszeroshot}.
This design enables an embodied “think-while-acting” paradigm, enhancing robustness, temporal continuity, and causal consistency in long-horizon manipulation tasks.

Our main contributions are summarized as follows:
\begin{itemize}
    \item We propose \textbf{HiF-VLA}, a VLA framework endowed with bidirectional spatio-temporal completion. By incorporating motion vectors as structured, low-dimensional temporal primitives, HiF-VLA explicitly expands the temporal receptive field, enabling temporally consistent and efficient action prediction while reducing redundancy.
    \item We propose a \textbf{hindsight-modulated joint expert} that unifies temporal and action representations within a unified space, enabling a “think-while-acting” paradigm for causally consistent and temporally coherent long-horizon motion generation.
    \item Extensive experiments demonstrate that our approach achieves substantial performance gains on widely adopted long-horizon benchmarks, while exhibiting strong temporal scalability and high inference efficiency.
\end{itemize}

\section{Related Work}\label{sec:related_work}

\textbf{Vision-Language-Action Models.} VLA models learn end-to-end mappings from language instructions and visual observations to low-level actions. Prior works can be broadly grouped by their action policies. Diffusion-based methods such as RDT-1B~\citep{liu2024rdt}, CogACT~\citep{li2024cogact}, and DexVLA~\citep{wen2025dexvla} employ iterative denoising to synthesize continuous control trajectories; autoregressive approaches like RT-2~\citep{zitkovich2023rt} and OpenVLA~\cite{kim2024openvla} discretize continuous actions into tokens and predict them sequentially using VLM backbones; regression-oriented systems such as OpenVLA-OFT~\citep{kim2025fine}, VLA-Adapter~\citep{wang2025vla} and SF~\citep{li2025spatial} adopt bidirectional attention to directly learn continuous actions via L1 regression. However, most VLA models do not explicitly model temporal dependencies. They often neglect the importance of historical information and reasoning capabilities through VLMs, both of which have been shown to substantially improve performance in long-horizon video tasks~\citep{chen2024diffusion,guo2025long,song2025history}.

\noindent\textbf{Temporal Modeling and Inference in Robotics.} Temporal modeling and foresight reasoning have been extensively studied in long video understanding and generation, but remain relatively underexplored in robotic manipulation. Existing approaches largely focus on one-sided reasoning. One line of work incorporates historical frames as visual prompts within the VLM input context, as in TraceVLA~\citep{zhengtracevla}, Octo~\citep{team2024octo}, GR-2~\citep{cheang2024gr}, RoboVLMs~\citep{liu2025towards}. Another emphasizes foresight by predicting future visual subgoals to guide action generation, such as CoT-VLA~\citep{zhao2025cot}, Seer~\citep{tianpredictive}, UniVLA~\citep{wang2025unified}, and UP-VLA~\citep{zhang2025up}, which typically condition an inverse dynamics model (IDM) on predicted future subgoals to infer actions. However, frame-level historical encoding and pixel-level future prediction introduce substantial computational overhead and redundancy, while failing to capture the importance of bidirectional temporal information. To address these, we propose HiF-VLA, which efficiently unifies hindsight, insight, and foresight within a single framework, enabling temporally coordinated decision-making for action generation.

\section{Method}

\subsection{Preliminary}
\begin{figure*}[ht]
    \centering
    \includegraphics[width=1\linewidth]{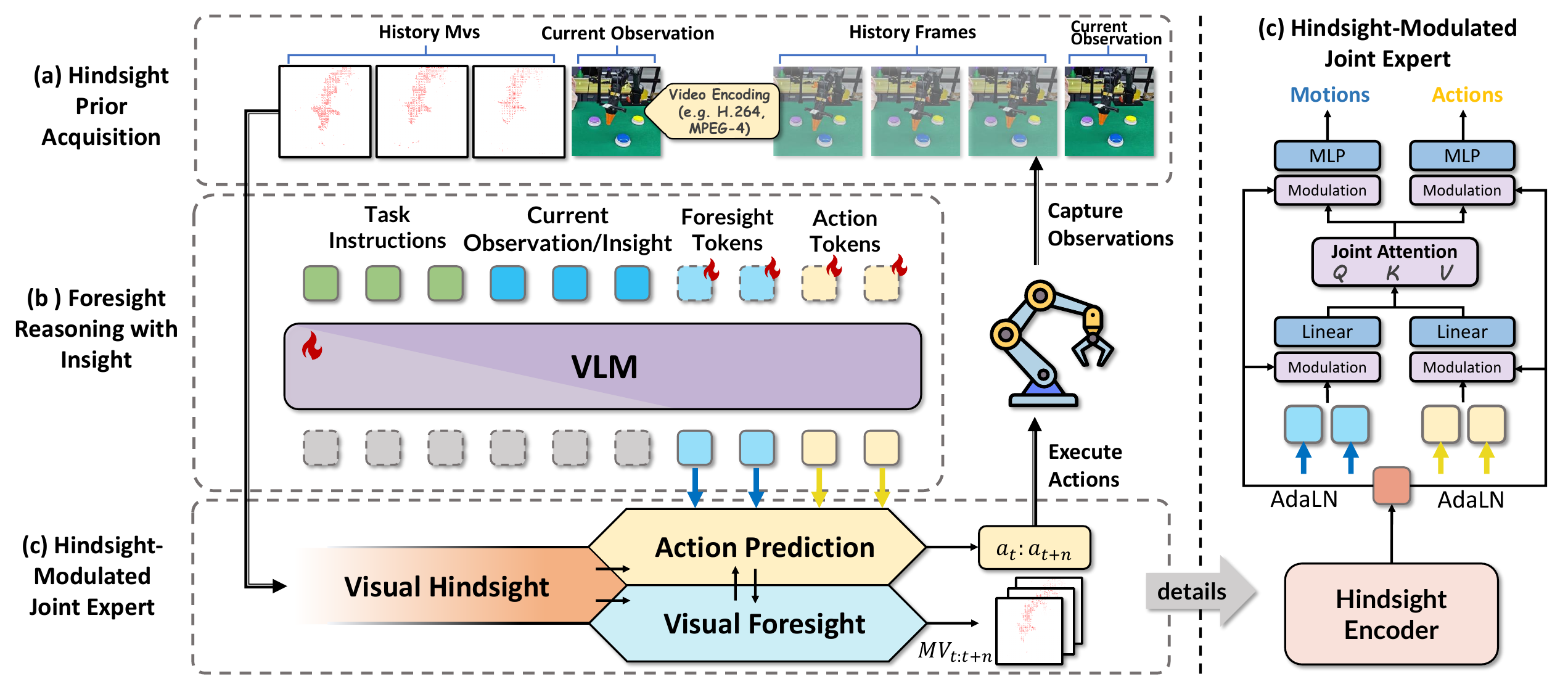}
    \caption{HiF-VLA Pipeline. (a) In Hindsight Prior Acquisition (see~\cref{sec:hindsight}), HiF-VLA encodes dense historical frame sequences into compact Motion Vector (MV) streams, forming structured hindsight primitives that capture temporal dynamics without pixel redundancy. (b) In Foresight Reasoning with Insight (see~\cref{sec:foresight}), the VLM interprets the task instruction and current observation to infer plausible foresight motions and corresponding latent action tokens. (c) Finally, the Hindsight-Modulated Joint Expert (see~\cref{sec:joint}) fuses hindsight, foresight, and action representations within a unified latent space, producing temporally consistent and causally coherent action predictions.}
    \label{fig:primary}
    \vspace{-8pt}
\end{figure*}

We begin by defining the setting and notation for VLA reasoning. Robot expert demonstrations are denoted as $D_r={(l,a_{1\ldots T},o_{1\ldots T} )}$. At each time step $t$, the model receives an observation $o_t$ and a textual instruction $l$, and the objective is to predict an action chunk $a_{t:t+n}$ over a horizon of length $n$. A vanilla VLA $P_\theta$~\citep{kim2025fine} builds on vision-language models (VLMs), learning from diverse manipulation demonstrations to transfer vision-language understanding and generation capabilities to embodied scenarios. Formally: 
\begin{equation}
     \tilde{a}_{t:t+n} \sim P_\theta\big(a_{t:t+n}\mid o_t, l\big). 
\end{equation}
Our key insight is to expand the temporal receptive field available prior to action execution by compensating sparse visual observations. To this end, we propose HiF-VLA, as illustrated in \cref{fig:primary}, a unified framework built upon a vanilla VLA architecture, which incorporates compact historical information prior $m_{t-h:t}^{his}$ of length $h$ and additionally predicts future motion $m_{t:t+n}$ conditioned on the current insight. Formally, the inference process can be expressed as:
\begin{equation}
(\tilde{a}_{t:t+n},\tilde{m}_{t:t+n}){\sim}P_\theta'(a_{t:t+n},m_{t:t+n}|o_t,l,m_{t-h:t}^{his}).
\end{equation}
During training, we augment the input with historical information and jointly predict both the future $n$-step motion and the corresponding actions. During inference, motion decoding is optional and can be omitted depending on downstream task requirements.

\subsection{Hindsight Prior Acquisition}\label{sec:hindsight}
Relying solely on instantaneous visual observations for action decision-making often fails to provide stable perception, especially under challenging conditions such as occlusion or repeated executions. Hence, capturing consistent dynamic patterns of the manipulator from historical states is critical. Existing approaches~\citep{cheang2024gr,tianpredictive,team2024octo} typically maintain a sliding window of past observations and concatenate them with the current observation to form a global representation. However, this frame-stacking strategy introduces significant redundancy: the high similarity between adjacent frames makes it difficult for the model to focus on task-relevant dynamics, potentially dispersing its attention.

To address this, we introduce a compressed historical prior: Motion Vectors (MVs)~\citep{jin2024video}. In video codec standards like H.264~\citep{wiegand2003overview} and MPEG-4~\citep{le1991mpeg}, MVs predict the displacements of macroblocks between adjacent frames to avoid pixel-wise redundancy. Importantly, MVs are not a coarse approximation: combined with keyframes, they enable near-lossless video reconstruction and maintain high compression. This property provides a natural, efficient, and faithful solution for capturing historical dynamics. Formally, let $(x,y)$ denote the position of a macroblock in an image (size $H\times W\times 3$). The motion vector is defined as:
\begin{equation}
    MV_{t-1:t}(x,y)=(x_t-x_{t-1},y_t-y_{t-1}),
\end{equation}
where $(x_t,y_t )$ and $(x_{t-1} ,y_{t-1})$ denote the positions of the macroblock in consecutive frames 
$o_t$ and $o_{t-1}$. We adopt MPEG-4 to extract keyframes and motion information, regarding the current observation $o_t$ at time step $t$ as the keyframe, while maintaining a historical window of length $m$, forming GOP (Group of Pictures) units, $GOP=[MV_{t-m:t-m+1},...,MV_{t-2:t-1},MV_{t-1:t},o_t]$.
Here, MVs follow the MPEG-4 16$\times$16 macroblock layout and are represented as a tensor of size $h\times(H//16)\times(W//16)\times2$, compactly encoding the historical motion trajectories of entities in the scene. Compared to raw frames, this representation significantly reduces redundancy while retaining task-relevant dynamics, thereby providing structured spatiotemporal priors to the decision layer. 

Then, we adopt a lightweight ViT-based~\citep{dosovitskiy2020image} hindsight encoder, combined with shallow 3D convolutions, to encode hindsight motions into compact hindsight tokens $M_h \in \mathbb{R}^{K_h \times d}$.

\subsection{Foresight Reasoning with Insight}\label{sec:foresight}
Beyond historical cues, foresight is equally critical for robot tasks. It allows the agent to evaluate the consequences of its actions before execution, promoting more reliable action decision-making. Existing approaches, such as CoT-VLA~\citep{zhao2025cot} and UP-VLA~\citep{zhang2025up}, 
achieve visual CoT reasoning by generating future visual subgoals. However, these approaches depend on pixel-level predictions prone to local distortions and semantic drift. The resulting dense and redundant subgoals obscure task-relevant dynamics, while the reliance on discrete frame prediction (without continuous temporal modeling) further undermines temporal consistency in complex environments.

To address these limitations, we propose an efficient foresight reasoning mechanism guided by current insight, as shown in \cref{fig:primary}(b). Instead of predicting raw future pixels, we employ more general and structured MVs as spatiotemporal targets for future action execution. MVs compactly encode the trajectory evolution of the manipulator within the scene, effectively reducing pixel redundancy while providing a structured spatial prior.

Specifically, we introduce $K_f$ learnable foresight query tokens $\{q_1^f,q_2^f,...,q_{K_f}^{f}\}$, together with $K_a$ empty action tokens $\{q_1^a,q_2^a,...,q_{K_a}^a\}$, into the VLM embedding space. These tokens are concatenated with the original inputs (including the task instruction $l$, the current observation $o_t$), and then fed into VLM $F_\theta$, enabling parallel reasoning over continuous visual dynamics and action generation, resulting in foresight motion tokens $M_f\in\mathbb{R}^{K_f\times d}$ and action latent tokens $A_f\in\mathbb{R}^{K_a\times d}$. Formally, the reasoning process can be expressed as: $(M_f,A_f)=F_\theta(o_t,l)$. 

We aim for the model to reason about visual foresight and action execution in parallel, and subsequently integrate and refine information from these distinct reasoning streams. This design enriches the diversity of the VLM's internal thought process and unlocks the potential of parallel reasoning.

\subsection{Hindsight-Modulated Joint Expert}\label{sec:joint}
Conventionally, the action expert receives high-level representations from the VLM and translates them into low-level control signals. However, predicting actions in isolation lacks reasoning about the future dynamics. Motion, as the physical manifestation of actions in the visual space, provides complementary information: jointly predicting both motion and actions enables VLA models to better align semantic understanding with underlying dynamics.

To achieve such synergistic reasoning, we propose the Hindsight-Modulated Joint Expert, which jointly models action and motion as two complementary streams within a shared temporal latent space and where historical information is introduced as a conditional prior to guide future inference and suppress erroneous replay of historical trajectories. Importantly, we integrate historical motion priors as adaptive temporal conditions, rather than embedding them directly into the VLM input. Injecting extra history motion into the VLM risks disrupting the established alignment between visual and language modalities~\citep{zhang2025ta} (see \cref{fig:his_position}). Instead, historical motion tokens $M_h$ are encoded into a compact conditional representation and modulate the joint reasoning process via Adaptive Layer Normalization (AdaLN) conditioning, where layered modulation and regularization jointly constrain future motion-action patterns.

Formally, we define three types of sequence representations: hindsight motion tokens $M_h\in\mathbb{R}^{K_h\times d}$ (used only as conditional input), foresight motion latent tokens $M_f\in\mathbb{R}^{K_f\times d}$, and action tokens $A_f\in\mathbb{R}^{K_a\times d}$. As shown in \cref{fig:primary}(c), the foresight motion tokens and action tokens form two parallel streams, which interact via cross-stream joint attention while retaining separate FFNs to ensure complementary yet disentangled representations. The hindsight tokens $M_h$ are projected through a linear layer to obtain the conditioning vectors $h_c$, which is then injected into each joint expert module via AdaLN to modulate both foresight and action representations $z\in \{M_f,A_f\}$:
\begin{equation}
\tilde{M}_f,\tilde{A}_f=\mathrm{Joint}\mathrm{Expert}(M_f,A_f\mid h_c),
\end{equation}
\begin{equation}
\mathrm{AdaLN}(z;h_c)=\gamma(h_c)\cdot\frac{z-\mu(z)}{\sigma(z)}+\beta(h_c),
\end{equation}
where $\sigma(z)$ and $\mu(z)$ are the mean and standard deviation of $z$, $\gamma(h_c)$ and $\beta(h_c)$ are modulation parameters from $h_c$. 
The Joint Attention employs non-causal self-attention over a sequence formed by concatenating $M_f$ and $A_f$, from which Q, K, and V are jointly projected.

Finally, the fused motion and action representations are projected through their respective heads to generate the future motions $\tilde{m}_{t:t+n}$ and the actions $\tilde{a}_{t:t+n}$. In general, the hindsight-modulated joint expert not only explicitly models the complementary relationship between action and motion, but also employs historical conditioning regularization to guide temporally consistent and physically plausible behaviors, thereby enhancing the stability and causal consistency of action.

\subsection{Unified HiF-VLA Model}
\textbf{Input Representation}. Following OpenVLA-OFT~\citep{kim2025fine}, we feed the current observation into both the DINOv2~\citep{oquab2023dinov2} and SigLIP~\citep{zhai2023sigmoid} visual encoders to obtain a hybrid visual embedding. At the same time, we initialize a set of learnable foresight motion queries and empty action tokens. These tokens are concatenated with the instruction embedding and the current observation embedding to form the multi-modal input sequence. In parallel, the historical motion sequence is compressed via spatial-temporal convolution, aggregated by a ViT encoder, and projected into the latent space of the joint expert as the hindsight prior. 

\noindent\textbf{Parallel Decoding and Joint Expert}. The VLM adopts a non-causal attention mask, enabling joint prediction of future motion latents and action latents. The three streams of tokens (hindsight, foresight, predicted action) are then fused via the Hindsight-Modulated Joint Expert.

\noindent\textbf{Training Objective}. We now describe in detail how we train the model to predict both actions and motion. To ensure both action and motion predictions remain well-calibrated, we define two L1-loss objectives:
\begin{equation}
\mathcal{L}_{MV}=\frac{1}{n}\sum_{j=1}^n | m_{t+j}-\tilde{m}_{t+j}|,
\mathcal{L}_{A}=\frac{1}{n}\sum_{j=1}^n |a_{t+j}-\tilde{a}_{t+j} |.
\end{equation}

The overall loss combines the two objectives:
\begin{equation}\mathcal{L}_{all}=\mathcal{L}_A+\lambda\cdot\mathcal{L}_{MV},\end{equation}
where $\lambda$ is a balancing factor between action accuracy and motion reconstruction quality, set to 0.01.

\begin{table*}[ht]
\centering

\begingroup
\small
\setlength{\tabcolsep}{3pt}
\renewcommand{\arraystretch}{1}

\begin{tabularx}{\linewidth}{
@{} p{2.7cm} p{1cm}
*{10}{Y}
@{}
}
\toprule
\textbf{Method} & 
\textbf{\footnotesize \makecell[l]{ Avg.\\SR}} &
\footnotesize \makecell[c]{Put soup\\and box\\in basket} &
\footnotesize \makecell[c]{Put box\\and butter\\in basket} &
\footnotesize \makecell[c]{Turn on\\stove and\\put pot} &
\footnotesize \makecell[c]{Put bowl\\in drawer\\and close} &
\footnotesize \makecell[c]{Put mugs\\on left\\and right\\plates} &
\footnotesize \makecell[c]{Pick book\\and place\\it in back} &
\footnotesize \footnotesize \makecell[c]{Put mug\\on plate,\\ pudding\\right} &
\footnotesize \makecell[c]{Put soup\\and sauce\\in basket} &
\footnotesize \makecell[c]{Put both\\pots on\\stove} &
\footnotesize \makecell[c]{Put mug \\in\\ \scriptsize microwave\\and close} \\
\midrule
\multicolumn{11}{l}{\textit{Policy inputs: third-view image, language instruction}} \\
\midrule
OpenVLA~\citep{kim2024openvla} & 54.0 & 35.0 & 95.0 & 65.0 & 45.0 & 40.0 & 80.0 & 60.0 & 45.0 & 20.0 & 55.0 \\
UniVLA*~\citep{li2025unified} & 63.0 & 64.0 & 82.0 & 76.0 & 96.0 & 58.0 & {98.0} & 24.0 & 74.0 & 32.0 & 26.0 \\
MemoryVLA~\citep{shi2025memoryvla} & 93.4 & 92.0 & {96.0} & 96.0 & \textbf{100} & \textbf{100} & \textbf{100} & \textbf{96.0} & 96.0 & 62.0 & \textbf{{96.0}} \\
\footnotesize {OpenVLA-OFT}*~\citep{kim2025fine} & 91.0 & 82.0 & {96.0} & 96.0 & 94.0 & 90.0 & 96.0 & {92.0} & \textbf{100} & 70.0 & 94.0 \\
\rowcolor{gray!15}
\textbf{HiF-VLA(Ours)} & {\textbf{94.4}} & {\textbf{94.0}} & \textbf{98.0} & \textbf{100} & \textbf{100} & 94.0 & \textbf{100} & 90.0 & {98.0} & {\textbf{76.0}} & 94.0 \\
\midrule
\multicolumn{11}{l}{\textit{Policy inputs: third-person image, wrist-view image, language instruction}} \\
\midrule
Seer (scratch)~\citep{tianpredictive} & 78.7 & 80.0 & 90.0 & 91.7 & 81.7 & 85.0 & 65.0 & 86.7 & 88.3 & 51.7 & 66.7 \\
Seer~\citep{tianpredictive} & 87.7 & 91.7 & 90.0 & 98.3 & \textbf{100} & 91.7 & 93.3 & 85.0 & 88.3 & 61.7 & 71.7 \\

UniVLA*~\citep{li2025unified} & 90.0 & \textbf{100} & 92.0 & 94.0 & {98.0} & 86.0 & \textbf{100} & 80.0 & \textbf{100} & 70.0 & 82.0 \\

\footnotesize {OpenVLA-OFT}*~\citep{kim2025fine} & 94.0 & 90.0 & \textbf{98.0} & {98.0} & {98.0} & {96.0} & \textbf{100} & {92.0} & \textbf{100} & 72.0 & {96.0} \\

\rowcolor{gray!15}
\textbf{HiF-VLA(Ours)} & \textbf{96.4} & 88.0 & \textbf{98.0} & \textbf{100} & \textbf{100} & \textbf{100} & \textbf{100} & \textbf{96.0} & \textbf{100} & \textbf{82.0} & \textbf{100} \\

\bottomrule
\end{tabularx}
\caption{Performance comparison on the LIBERO-Long benchmark. We report the average success rate (\%) across 10 tasks with ``\textbf{Avg. SR}".
Results marked with ``$*$'' were reproduced using the official open-source code.
\textbf{Bold} indicates the best performance.}
\label{tab:libero_long}
\vspace{-8pt}
\endgroup
\end{table*}

\section{Experiments}
In this section, we design experiments to address the following research questions (\textbf{RQs}):

\noindent\textbf{RQ1}: How does HiF-VLA perform compared to SOTA methods on challenging long-horizon benchmarks?

\noindent\textbf{RQ2}: Does HiF-VLA effectively mitigate the redundancy and inefficiency issues in conventional approaches?

\noindent\textbf{RQ3}: How well does HiF-VLA maintain inference scalability as the temporal horizon increases?

\noindent\textbf{RQ4}: Through ablation studies, how do different components contribute to HiF-VLA’s overall performance?

\noindent\textbf{RQ5}: Can HiF-VLA successfully handle long-horizon tasks on real-world robotic platforms?

\subsection{Overall Performance}

\noindent\textbf{Experimental Setups.} We evaluate HiF-VLA on two long-horizon benchmarks.
LIBERO-Long~\citep{liu2023libero} comprises ten multi-subgoal manipulation tasks across diverse scenes.
CALVIN ABC-D~\citep{mees2022calvin} includes four indoor environments (A-D); policies are trained on A-C and evaluated on the unseen D to assess generalization on consecutive tasks.
All experiments are conducted under two settings following~\citep{kim2025fine}: a third-view setup using the primary camera, or a multi-view setup using both the primary and wrist cameras.

\noindent\textbf{Implementation Details.}
We adopt Prismatic-7B~\citep{karamcheti2024prismatic} as the VLM backbone, and initialize it with weights from OpenVLA~\citep{kim2024openvla}, which were pretrained on OXE~\citep{o2024open}. All other modules are randomly initialized. Training is performed on 8 NVIDIA A100 GPUs with a global batch size of 64. A fixed temporal chunk of \textbf{n=8} is adopted for both action and foresight modeling, while the hindsight window is variable (\textbf{default 8}). Fine-tuning is conducted for 150k steps on LIBERO and 80k on CALVIN. The main baseline is OpenVLA-OFT~\citep{kim2025fine}, which shares the same pretrained initialization. Additional comparisons include Seer~\citep{tianpredictive}, VPP~\citep{huvideo}, and $\pi_0$~\citep{black2024pi_0}, etc.

\noindent\textbf{Result Analysis.} 
\textbf{1) LIBERO-Long:} As shown in \cref{tab:libero_long}, we present the detailed performance of HiF-VLA across 10 tasks in the LIBERO-Long benchmark. We evaluate both third-view and multi-view inputs over 500 trials. Compared to the baseline third-view method, our approach achieves a 94.4\% success rate, representing a 3.4\% absolute improvement. Notably, our third-view variant performs on par with multi-view baselines, underscoring HiF-VLA’s robust temporal reasoning capability. Moreover, our method achieves a 96.4\% success rate under the multi-view setting, consistently outperforming other state-of-the-art VLA models.
\textbf{2) CALVIN-ABC-D}: 
We further evaluate the generalization capability of HiF-VLA, which was trained on the ABC dataset and evaluated in the D environment. As shown in \cref{tab:third_multi_view_comparison}, our method surpasses the baseline by 0.25 in terms of the average task length metric and achieves superior performance under both third-view and multi-view settings. These results clearly answer \textbf{RQ1}. This improvement can be attributed to the bidirectional temporal perception and reasoning architecture incorporated in our model, which enables a more effective understanding of long-term action dependencies and consequently leads to enhanced adaptability and robustness in complex long-horizon tasks.

\begin{table}[t]
\small
\centering
\setlength{\tabcolsep}{3pt}
\renewcommand{\arraystretch}{1.1}
\begin{tabular}{lccccc|c}
\toprule
 \textbf{Method} & 1 & 2 & 3 & 4 & 5 & \textbf{Avg. Len.} $\uparrow$ \\
\midrule
\multicolumn{7}{l}{\textit{Policy inputs: third-view image, instruction}}\\
\midrule
SuSIE \cite{blackzero} & 87.0 & 69.0 & 49.0 & 38.0 & 26.0 & 2.69 \\
 OpenVLA \cite{kim2024openvla} & 91.3 & 77.8 & 62.0 & 52.1 & 43.5 & 3.27 \\
 CLOVER \cite{bu2024closed} & \textbf{{96.0}} & 83.5 & 70.8 & 57.5 & 45.4 & 3.53 \\
 VPP \cite{huvideo} &90.9&81.5&71.3&62.0&51.8&3.58\\
 $\pi_0$ \cite{black2024pi_0} &93.7&83.2&74.0&62.9&51.0&3.65\\
 UniVLA \cite{bu2025univla} & 95.5 & 85.8 & 74.8 & 66.9 & 56.5 & 3.80 \\
\rowcolor{gray!10}
 \textbf{HiF-VLA(Ours)} & 93.5 & \textbf{87.4} & \textbf{81.4} & \textbf{75.9} & \textbf{69.4} & \textbf{4.08} \\
\midrule
\multicolumn{7}{l}{\textit{Policy inputs: third-person and wrist-view image, instruction}} \\
\midrule
 GR-1 \cite{wuunleashing} & 85.4 & 71.2 & 59.6 & 49.7 & 40.1 & 3.06 \\
 Vidman \cite{wen2024vidman} & 91.5 & 76.4 & 68.2 & 59.2 & 46.7 & 3.42 \\
 $\pi_0$ \cite{black2024pi_0} & 93.8 & 85.0 & 76.7 & 68.1 & 59.9 & 3.92 \\
 UP-VLA \cite{zhang2025up} & 92.8 & 86.5 & 81.5 & 76.9 & 69.9 & 4.08 \\
 OpenVLA-OFT \cite{kim2025fine} & 96.3 & 89.1 & 82.4 & 75.8 & 66.5 & 4.10 \\
 RoboVLMs \cite{liu2025towards} &98.0&93.6 &85.4&77.8&70.4&4.25 \\
 Seer \cite{tianpredictive} & 96.3 & 91.6 & 86.1 & 80.3 & 74.0 & 4.28 \\
VPP \cite{huvideo} & 96.5 & 90.9 & 86.6 & \textbf{82.0} & \textbf{76.9} & 4.33 \\
\rowcolor{gray!10}
\textbf{HiF-VLA(Ours)} & \textbf{98.5} & \textbf{94.1} & \textbf{88.1} & 81.4 & 73.1 & \textbf{4.35} \\
\bottomrule
\end{tabular}
\caption{Performance comparison on the CALVIN ABC-D benchmarks. We report the average number of successfully completed tasks across five consecutive instructions. \textbf{Bold} indicates the best performance.}
\label{tab:third_multi_view_comparison}
\vspace{-12pt}
\end{table}

\begin{figure*}[htbp]
    \centering
    
    \begin{subfigure}[b]{0.24\textwidth}  %
        \centering        \includegraphics[width=\textwidth]{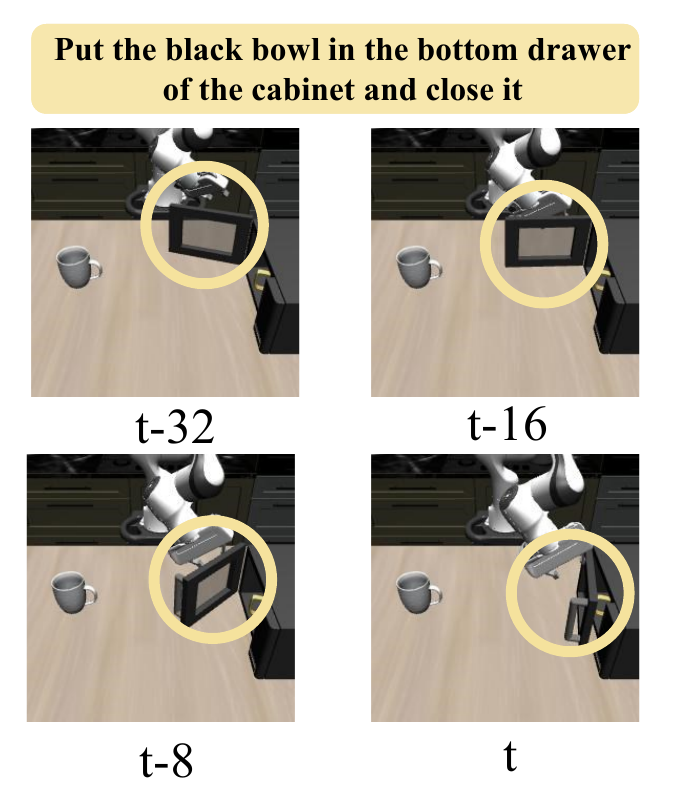}  %
        \caption{Historical Frame Example}
        \label{fig:hisfig-a}
    \end{subfigure}
    \begin{subfigure}[b]{0.355\textwidth}
        \centering        \includegraphics[width=\textwidth]{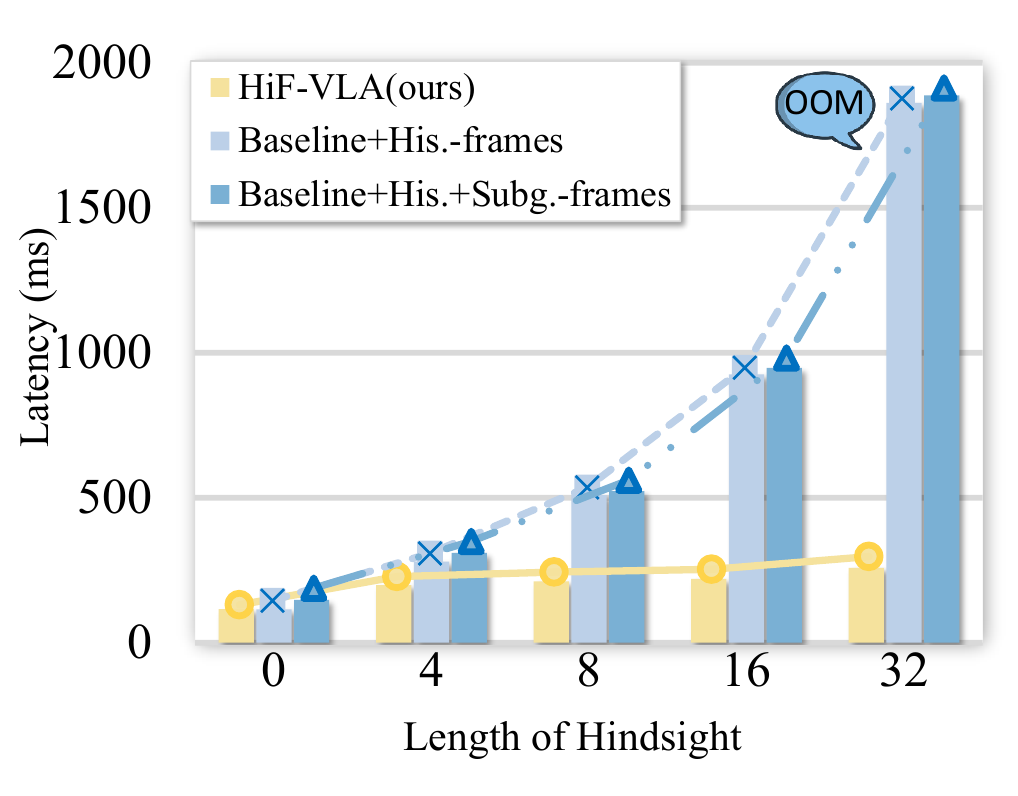}  %
        \caption{Inference Efficiency}
        \label{fig:hisfig-b}
    \end{subfigure}
    \begin{subfigure}[b]{0.355\textwidth}
        \centering        \includegraphics[width=\textwidth]{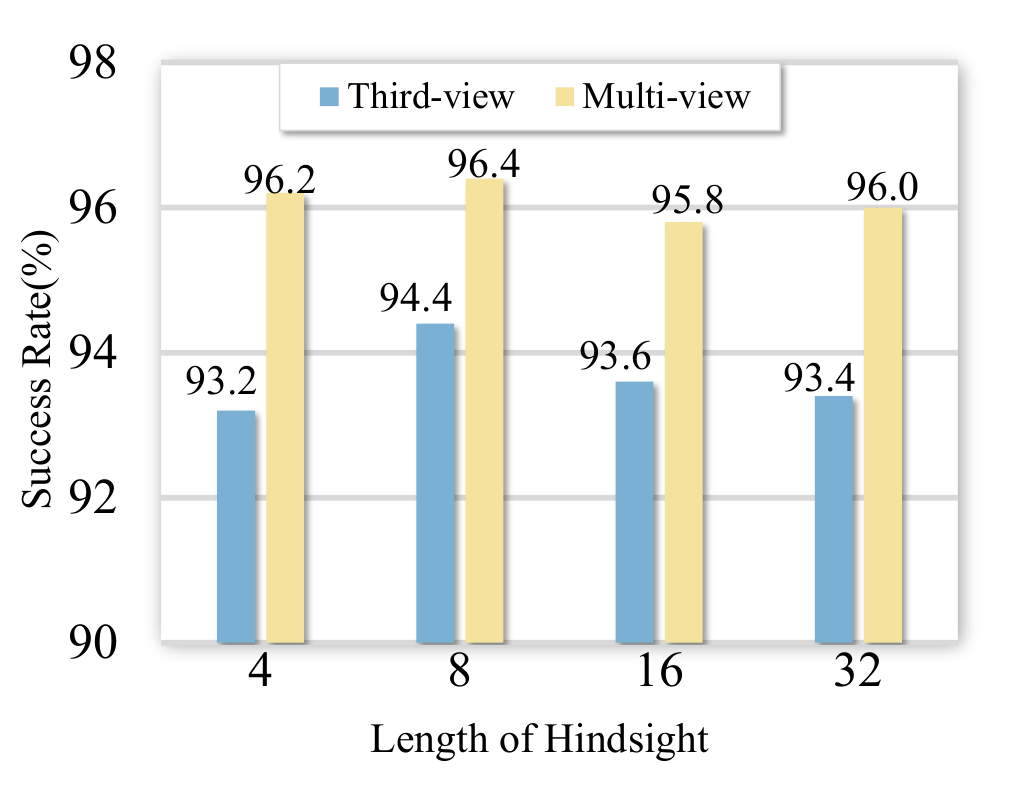}  %
        \caption{Effect of Hindsight}
        \label{fig:hisfig-c}
    \end{subfigure}
    \vspace{-3pt}
    \caption{Effect of hindsight length on performance and efficiency. (a) Example of historical frames. (b) HiF-VLA maintains low inference latency as hindsight length increases. (c) Performance of hindsight of different lengths in third-view and multi-view perspectives.}
    \label{fig:two_subfigs}
    \vspace{-8pt}
\end{figure*}

\subsection{Efficiency and Redundancy Analysis}

\noindent\textbf{Experimental Setup.} 
We conduct experiments exclusively on {LIBERO-Long} to evaluate the efficiency and redundancy aspects, and the results are presented in \cref{tab:efficient}. 
For the baseline (1), we adopt the method proposed by~\citep{kim2025fine}.  
(2) ``{+ Subgoal}'' and (4) ``{+ History Frames}'' correspond to variants that incorporate RGB-based prediction and historical information following~\citep{cheang2024gr,zhao2025cot}.  
(3) ``{+ Foresight (Ours)}'' and (5) ``{+ Hindsight (Ours)}'' denote variants that introduce Motion-based foresight and historical information, respectively.  
(6) ``{+ Hindsight + Foresight (Ours)}'' integrates both types of Motion-based information simultaneously.  
Note that we used third-person input, a hindsight length of 4, and a batch size of 4 in all experiments.

\begin{table}
    \centering
    \renewcommand{\arraystretch}{1.1}
    \resizebox{\columnwidth}{!}{%
    \begin{tabular}{clcc|c}
    \toprule
        & \textbf{Methods}&  \makecell[c]{\textbf{Peak GPU}\\\textbf{Memory}\textbf{(GB)} \textbf{$\downarrow$}}& \makecell[c]{\textbf{Latency}\\\textbf{(ms) }\textbf{$\downarrow$}} & \makecell[c]{\textbf{Avg.}\\\textbf{SR$\uparrow$}}\\
    \midrule
     (1) & Baseline  & 30.8 (1.00$\times$) & 72.9 (1.00$\times$) & 91.0 \\
      (2) &+ Subgoal  & 38.2  (1.24$\times$) &115.9 (1.59$\times$)  & 91.8 \\
       \rowcolor{gray!10}
       (3) &+ Foresight (Ours) & 31.8 (1.03$\times$) & 82.7 (1.13$\times$) &92.2 \\
    \midrule
     (4) & + History frames  & 63.6 (2.06$\times$) & 229.5 (3.15$\times$) &  90.4\\
       \rowcolor{gray!10}
      (5)& + Hindsight (Ours) & 31.4 (1.02$\times$) & 117.7 (1.61$\times$) & 92.2 \\
    \midrule
    \rowcolor{gray!10}
      (6) & + Hindsight + Foresight (Ours) &32.2 (1.05$\times$)  & 121.6 (1.67$\times$) & 93.2\\
    \bottomrule
    \end{tabular}
    }
    \caption{Performance comparison between multi-frame baselines variants and HiF-VLA variants. Peak GPU memory (GB) during training and inference latency (ms) are reported. The history and hindsight lengths are fixed to 4 for a fair and computationally feasible comparison across all methods.}
    \label{tab:efficient}
    \vspace{-15pt}
\end{table}

\noindent\textbf{Result Analysis.} As shown in \cref{tab:efficient}(2) and (3), incorporating subgoal or foresight prediction improves task success rates; however, subgoal-based methods incur substantial latency overhead (1.59$\times$ slower than the baseline). In contrast,  our foresight head introduces negligible additional cost (0.13$\times$ latency and 0.03$\times$ GPU Memory). Moreover, as illustrated in \cref{tab:efficient}(4), dense multi-frame inputs significantly slow down inference (229.5 ms, 3.15$\times$ slower than the baseline) and even degrade performance, suggesting that redundant pixel information dilutes task-relevant temporal cues and may cause overfitting to visually irrelevant details. 
To address this, HiF-VLA replaces dense RGB inputs with compact motion representations, allowing the model to focus on dynamic and task-relevant cues, thereby improving both efficiency and accuracy. Furthermore, unified integration of hindsight and foresight further enhances overall performance. 
These results highlight HiF-VLA’s clear advantages in reducing redundancy and improving inference efficiency, effectively addressing \textbf{RQ2}. 

\begin{figure}[htbp]
    \centering
    \vspace{-7pt}
    \begin{subfigure}[b]{0.19\textwidth}  %
        \centering
        \includegraphics[width=\textwidth]{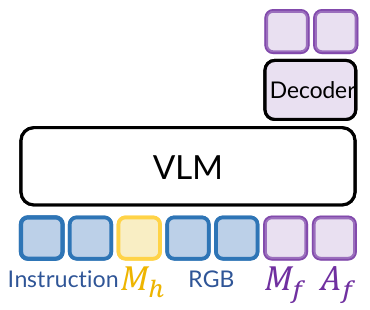}  %
        \caption{Injecting VLM}
        \label{fig:subfig-a}
    \end{subfigure}
    \begin{subfigure}[b]{0.19\textwidth}
        \centering
        \includegraphics[width=\textwidth]{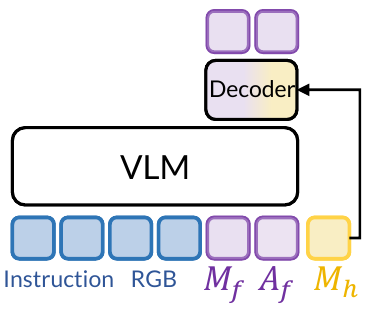}  %
        \caption{Conditioning Decoder}
        \label{fig:subfig-b}
    \end{subfigure}
    \begin{subfigure}[b]{0.08\textwidth}
        \centering
        \includegraphics[width=\textwidth]{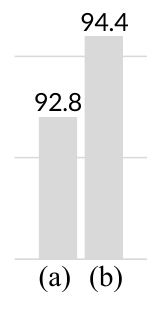}  %
        \caption{Results}
        \label{fig:subfig-c}
    \end{subfigure}
    \caption{Performance comparison on different hindsight embedding locations. (a) represents direct injection into the VLM, and (b) represents conditional embedding as an expert decoder. (c) shows the performance of both on LIBERO-Long. $M_h$ denotes the hindsight tokens, $M_f$ represents the foresight tokens and $A_f$ is the action tokens.}
    \label{fig:his_position}
    \vspace{-11pt}
\end{figure}

\subsection{Inference Scalability}

\noindent\textbf{Experimental Setup.} We evaluate the inference efficiency of HiF-VLA against two multi-frame baselines~\citep{kim2025fine}: (i) multi-frame extension with stacked past observations, and (ii) multi-frame history combined with single-frame subgoal prediction. For each model variant, we conduct 100 inference runs on an NVIDIA A100 GPU and report the average latency (defined as the time required to generate one action chunk) in the LIBERO-Long simulation.

\noindent\textbf{Results Analysis.} As shown in \cref{fig:hisfig-b}, the latency of multi-frame baselines increases almost linearly with the length of history, reflecting their high computational sensitivity to frame stacking. For instance, at a history length of 8, the baseline incurs over 4.5× higher latency compared to the vanilla VLA. In contrast, HiF-VLA maintains a consistently low computational overhead across all context lengths, with latency increasing only marginally as history grows, demonstrating superior scalability with respect to temporal context length, thereby directly addresses \textbf{RQ3}.

\begin{figure*}[ht]
    \centering
\includegraphics[width=1\linewidth]{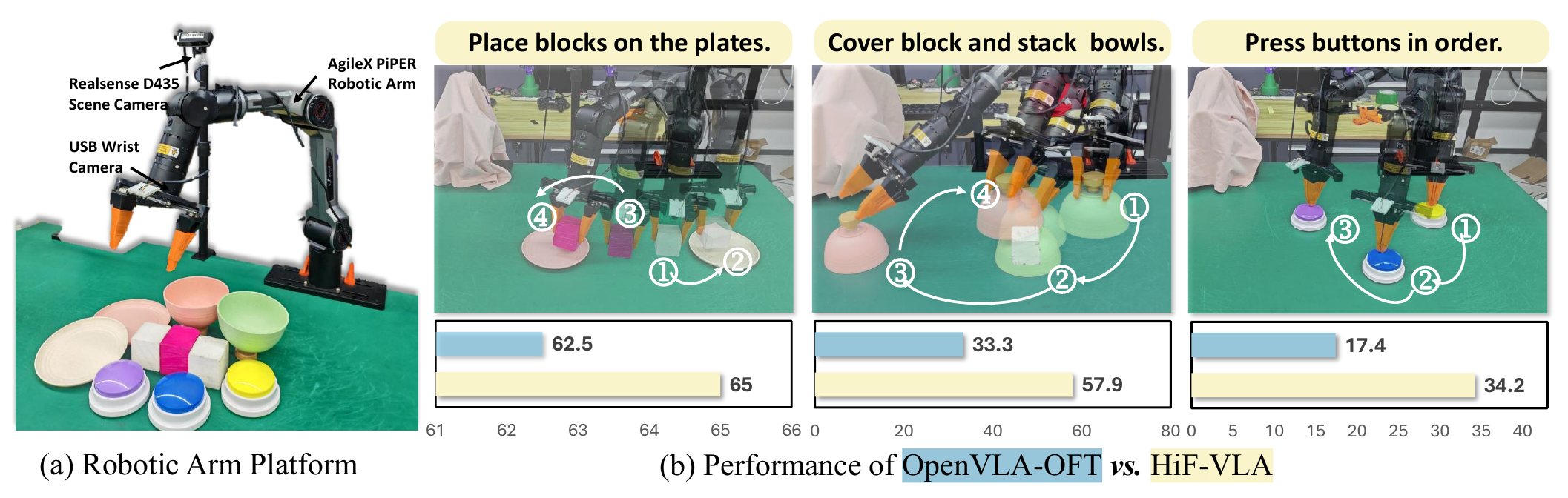}
\vspace{-19pt}
 \caption{Real-world long-horizon tasks. (a) We deploy our system on the AgileX Piper robotic arm equipped with an external scene camera (Intel RealSense D435) and a wrist-mounted camera. (b) We design three long-horizon tasks covering diverse primitives such as pick, put, cover, stack, and press, emphasizing temporal consistency in action generation.}
 \vspace{-9pt}
    \label{fig:realworld_rgb}
\end{figure*}

\subsection{Ablation Studies}
\noindent\textbf{Experimental Setup.} 
To address \textbf{RQ4} regarding the optimal integration of historical information, the effects of hindsight length and historical embedding positions are investigated on the LIBERO-Long.

\noindent\textbf{Hindsight Length.} We examine the impact of hindsight length in both third-view and multi-view settings. As illustrated in \cref{fig:hisfig-c}, the model achieves peak performance $94.4\%$ and $96.4\%$ when the hindsight length is set to 8. We believe, in the LIBERO-Long benchmark, most manipulation sequences exhibit moderate temporal dependencies, where 8 hindsight lengths are sufficient to capture meaningful causal cues without introducing redundant information.

\noindent\textbf{Hindsight Embedding Position.} We study how the embedding strategy of hindsight information affects model performance. Specifically, we compare two variants: (i) naïvely concatenating hindsight with language and vision as direct inputs to the VLM, and (ii) conditioning hindsight within the expert module, which is adopted in our implementation. As shown in \cref{fig:his_position}, the expert-conditioned embedding consistently achieves higher success rates than VLM-based embeddings. We attribute this to the fact that motion-based hindsight may interfere with the pretrained alignment between visual and language representations. 
In contrast, injecting hindsight at the decoding stage provides a more direct, residual-like path for motion information. This allows the low-level dynamics encoded in the hindsight to guide future action prediction without passing through, and potentially being corrupted by the VLM's semantic fusion layers.

\subsection{Real-world Experiments}

\noindent\textbf{Experiment setups.} To evaluate the effectiveness of our approach in real-world applications, we conduct real-world experiments using the AgileX Piper robot. As shown in ~\cref{fig:realworld_rgb}(a), a RealSense D435 camera captures the scene from a third-view, while an additional USB camera is mounted on the robot’s wrist for egocentric observations. We collect three long-horizon tasks, each with 100 demonstrations involving diverse manipulation primitives, including pick, place, stack, cover, and press.

\noindent\textbf{Real-world evaluation.}
For real-world environments, we train each model separately for every task and evaluate performance by averaging success rates over 20 trials per task. These long-horizon tasks require the model to maintain action consistency across stages, correctly associate time states. As shown in ~\cref{fig:realworld_rgb}(b), the baseline model (OpenVLA-OFT) performs poorly across these long-horizon tasks — for example, achieving only 17.4\% success on Press-Buttons-Order, often failing to complete required button presses. This is likely due to the minimal visual difference between pressed and unpressed states, making it difficult for the baseline to detect successful actuation. In contrast, HiF-VLA benefits from its broad temporal receptive field, enabling reliable detection of subtle state transitions and robust execution of long-horizon tasks, resulting in superior performance across real-world tasks.

\section{Conclusion}

This work introduces HiF-VLA, an efficient and unified framework for temporal perception and reasoning built upon low-dimensional, structured motion vectors. By integrating hindsight, insight, and foresight cues, HiF-VLA establishes a bidirectional temporal expansion over a sparse visual receptive field, enabling the robot to capture task-critical dynamics at minimal computational cost and thereby improving temporal consistency and causal coherence in long-horizon tasks. Across both simulated and real-world long-horizon manipulation benchmarks, HiF-VLA demonstrates strong performance. 
\textbf{Limitations.} The current motion representation remains dependent on estimation accuracy and may be sensitive to noise in highly dynamic scenes. 
We leave the exploration of large-scale pre-training on internet videos to enhance motion understanding and generation capabilities for future work.

\section*{Acknowledgements}

This work was supported by the National Science and Technology Major Project (Grant No. 2022ZD0208800) and the National Natural Science Foundation of China General Program (Grant No. 62573362).

{
    \small
    \bibliographystyle{ieeenat_fullname}
    \bibliography{main}
}

\clearpage
\setcounter{page}{1}
\maketitlesupplementary

\section{More Implementation Details}
\label{sec:rationale}
Beyond the SigLIP~\cite{zhai2023sigmoid} and DINOv2~\cite{oquab2023dinov2} image encoders and the Prismatic VLM~\cite{karamcheti2024prismatic} backbone described in the main text, we provide further additional implementation details for the two core modules used in HiF-VLA: the Hindsight Encoder and the Hindsight-Modulated Joint Expert. These details complement the high-level description given in the main text.

\subsection{Hindsight Encoder}
We employ a 4-layer Vision Transformer (ViT)~\cite{dosovitskiy2020image} as the hindsight motion encoder. The hindsight motion sequence of length $h$ is first partitioned into $(h//2)\times(H//2)\times(W//2)$ spatiotemporal blocks using a 3D convolution, which simultaneously reduces temporal redundancy and preserves local motion continuity. These blocks, together with an additional $[CLS]$ token that aggregates global temporal context, are fed into the Transformer. The resulting historical features are subsequently projected via a linear layer into the joint expert embedding space, ensuring dimensional compatibility with the foresight and action pathways. This design allows the hindsight encoder to provide structured historical priors that can effectively regularize downstream decision-making.

\subsection{Hindsight-Modulated Joint Expert}
The hindsight-modulated joint expert serves as the fusion module that integrates hindsight, foresight, and action representations. All tokens are projected into a shared embedding dimension of 1024, including: hindsight motion tokens from the hindsight encoder, foresight motion tokens and latent action tokens produced by the VLM backbone. Within this space, the hindsight tokens serve as adaptive conditioning signals that modulate both foresight and action streams via AdaLN~\cite{xu2019understanding} scaling and shifting. This conditioning mechanism allows the model to dynamically adjust its predictions based on temporally grounded historical cues, enabling causal alignment between past observations and future behavior. Positional information is provided through Rotary Positional Embedding (RoPE)~\cite{su2024roformer}, allowing efficient encoding of both spatial and temporal ordering. The joint expert consists of 6 Transformer layers, each capable of cross-stream attention, enabling the model to capture complementary relationships between predicted dynamics and actions.
\begin{figure}[t]
    \centering
    \includegraphics[width=1\linewidth]{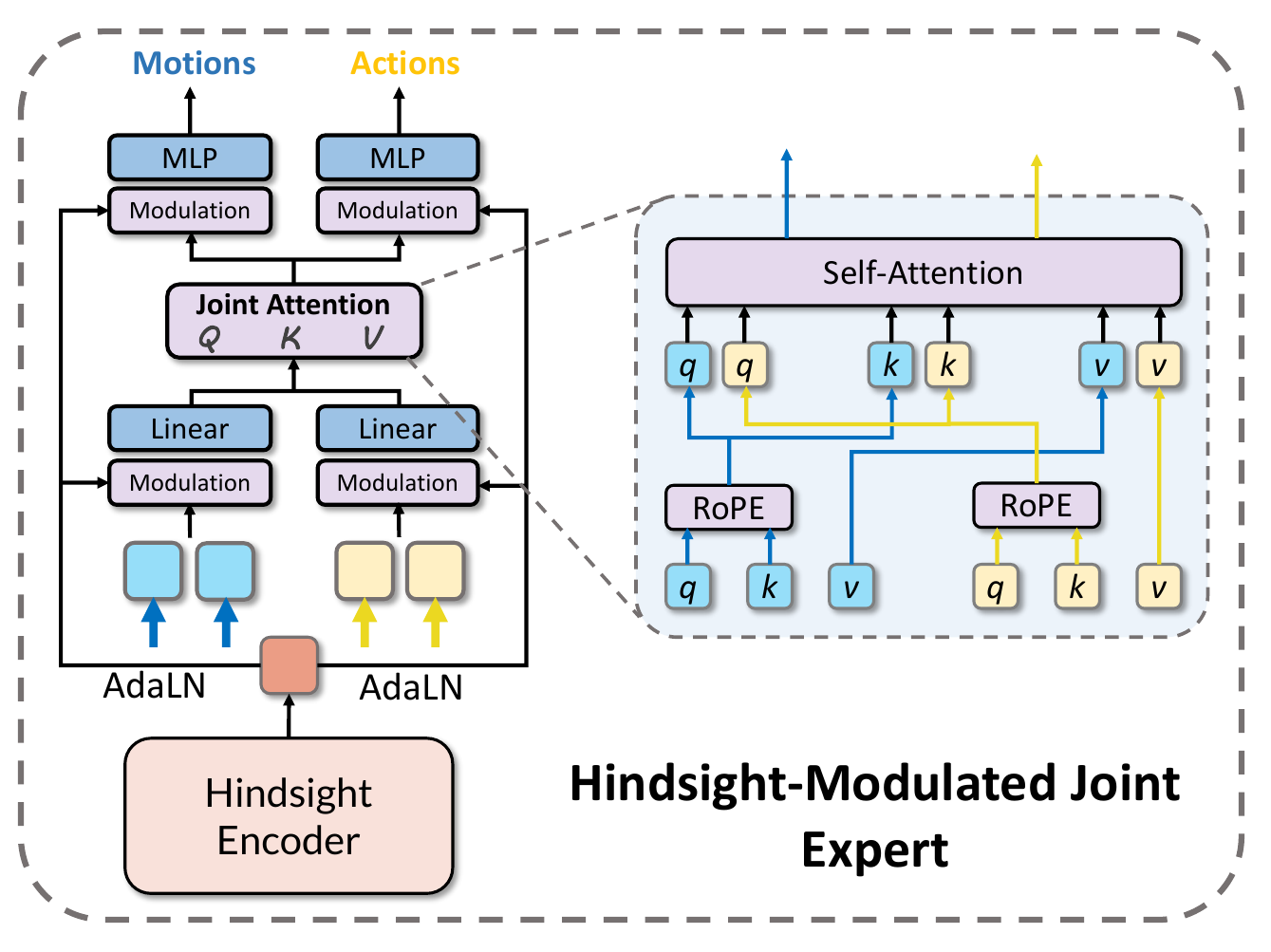}
    \caption{Architecture of the hindsight-modulated joint expert.}
    \label{fig:joint_expert}
\end{figure}

\begin{table*}[ht]
    \centering
    \begin{tabular}{cccccc}
        \toprule
      \textbf{Methods} &\textbf{LIBERO-Spatial}& \textbf{LIBERO-Object }&\textbf{LIBERO-Goal}& \textbf{LIBERO-Long}& \textbf{Average}\\
         \midrule
        TraceVLA~\cite{zhengtracevla}&84.6  &85.2  & 75.1 & 54.1 & 74.8\\
        Octo~\cite{team2024octo} & 78.9 & 85.7 & 84.6 & 51.1 & 75.1\\
        CoT-VLA~\cite{zhao2025cot} & 81.1 & 87.5 & 91.6 & 87.6 & 69.0\\
        SpatialVLA~\cite{li2025spatial} &88.2&89.9&78.6&55.5&78.1\\
        ThinkAct~\cite{huangthinkact}& 88.3 & 91.4 & 87.1 & 70.9 &84.4 \\
        Seer~\cite{tianpredictive} &-&-&-&87.7&87.7\\
        FlowVLA~\cite{zhong2025flowvla} & 93.2 & 95.0 & 91.6 & 72.6 & 88.1\\
        4D-VLA~\cite{zhang4d}&88.9&95.2&90.9&79.1&88.6\\
        DreamVLA~\cite{zhang2025dreamvla} &97.5  & 94.0 & 89.5 & 89.5 &92.6 \\
        CogACT~\cite{li2024cogact} & 97.2 & 08.0 & 90.2 & 88.8 & 93.2\\
        $\pi_0$~\cite{black2024pi_0} & 96.8 & 98.8 & 95.8 & 85.2 &94.2 \\
        GR00T N1~\cite{bjorck2025gr00t} &94.4&97.6&93.0&90.6&93.9\\
        UniVLA~\cite{bu2025univla} & 96.5 & 96.8 & 95.6 & 92.0 &95.2 \\
        MemoryVLA~\cite{shi2025memoryvla}& 98.4 & 98.4 & 96.4 &  93.4& 96.5\\
        OpenVLA-OFT~\cite{kim2025fine} & 97.6 & 98.4 & \textbf{97.9} & 94.5 &97.1 \\
        \rowcolor{gray!10}
        \textbf{HiF-VLA(ours)} & \textbf{98.8} & \textbf{99.4}  & 97.4 & \textbf{96.4} &\textbf{98.0} \\
        \bottomrule
    \end{tabular}
    \caption{Performance on the LIBERO benchmark. The table compares our method against a wide range of state-of-the-art approaches, with the best performance highlighted in \textbf{bold}.}
    \label{tab:libero_all}
\end{table*}

\section{Comparison with Video-Generation VLAs}
Compared to VLA approaches that rely on video generation~\cite{huvideo,wen2024vidman,cheang2024gr,wu2023unleashing}, our method differs fundamentally in how it models temporal dynamics. A large body of recent work~\cite{huvideo,wen2024vidman,cheang2024gr,wu2023unleashing} employs general-purpose video generative models to predict future frames, using these predictions either for inverse dynamics computation or as conditional representations to assist action policy generation. While these approaches have made substantial progress, they face some limitations when contrasted with HiF-VLA, which leverages sparse motion representations.

\begin{table}[t]
\centering
\begin{tabular}{c|lcc}
\toprule
&\textbf{Ablation} & \textbf{Parameters} & \textbf{SR} \\
\midrule
\multirow{4}{*}{(a)}&\multirow{4}{*}{\makecell[l]{Foresight Motion\\Loss Weight $\lambda$}}
& 0.1 & 94.4 \\
&& 0.05 & 95.2 \\
&& 0.01* & \textbf{96.4} \\
&& 0.001 & 95.6 \\
\midrule
\multirow{4}{*}{(b)}&\multirow{4}{*}{Joint Expert Depth}
& 2 & 95.2 \\
&& 4 & 95.6 \\
&& 6* & \textbf{96.4} \\
&& 8 & 95.2 \\
\midrule
\multirow{3}{*}{(c)}&\multirow{3}{*}{\makecell[l]{Length:\\(Hindsight, Foresight)}}
& (8,8)* & \textbf{96.4} \\
&& (8,16) & 94.6 \\
&& (16,16) & 95.2 \\
\bottomrule
\end{tabular}
\caption{Ablation on hyper-parameters. * denotes the submission setting. (multi-view)}
\label{tab:hyper_ablation}
\end{table}

Specifically, video-generation-based methods typically require multi-frame historical input and high-resolution future video synthesis, resulting in substantial computational overhead. Moreover, pixel-level prediction is prone to local artifacts and distortions, introducing uncontrollable noise. In contrast, HiF-VLA replaces video-level temporal modeling with low-dimensional, structured motion representations, enabling a more efficient and stable way to encode historical visual states and predict future dynamics. In addition, HiF-VLA, by jointly predicting foresight motion and action trajectories as two complementary information streams, allows the model to anticipate how the physical world may evolve while generating the corresponding actions—effectively enabling thinking while acting.

\section{More Experimental Results}

\subsection{Comprehensive Evaluation on the LIBERO Benchmark}
We report detailed evaluation results on all four suites of the LIBERO benchmark~\cite{liu2023libero} and compare our method against a broad set of baseline models, as summarized in \cref{tab:libero_all}. While achieving its greatest margin of superiority under the most challenging LIBERO-Long suite, HiF-VLA also delivers competitive or superior performance across the remaining three suites when compared to prior state-of-the-art approaches. As a result, HiF-VLA attains the best average performance over all four suites, demonstrating its robustness and general applicability across diverse task scenarios.

\subsection{More Hyper-parameters Ablation}

\textbf{The Impact of Weight Factor $\lambda$.} The hyperparameter $\lambda$ balances the contributions between foresight motion prediction and action prediction. To systematically investigate this trade-off, we evaluate a range of $\lambda$ values. Our analysis reveals that an appropriate weighting allows the motion prediction to effectively support the model's reasoning, thereby enhancing planning capabilities; however, an unsuitable value destabilizes the VLA architecture and impedes policy generation. Our results in \cref{tab:hyper_ablation}(a) identify an optimal performance point at $\lambda=0.01$. This key finding indicates that a modest, well-calibrated contribution from the motion-prediction branch is essential for achieving balanced and robust model behavior.

\textbf{Depth of the Joint Expert.} We investigated the optimal interaction depth within the Joint Expert by performing an ablation study over depths of $\{2, 4, 6, 8\}$ (see \cref{tab:hyper_ablation}(b)). We observed that a depth of 6 yields the best performance. Shallower networks facilitate the interaction between motion and action modalities, whereas increasing the depth further yields negligible performance gains, indicating depth is not highly critical beyond moderate depth.

\textbf{Foresight and Foresight Horizon.}  In \cref{tab:hyper_ablation} bottom, increasing foresight to $n=16$ hurts performance due to error accumulation in long-term prediction. In contrast, extending hindsight to 16 (keeping foresight at 16) improves results by providing richer context, supporting our choice to decouple hindsight and foresight lengths.

\subsection{More Variants Ablation}
\textbf{State replaces motion.} To better understand the role of motion representations, we conduct an ablation study replacing codec motion vectors (\textbf{M}) with robot proprioceptive states (\textbf{S}). This experiment is motivated by the observation that motion vectors might appear to primarily reflect robot arm movement. If this were the case, robot states—which explicitly encode the robot configuration—should provide similar information and lead to comparable performance. However, motion vectors describe inter-frame visual changes and therefore capture not only robot motion but also the visual consequences of interactions with the environment, such as object displacement, contact-induced motion, or changes in scene structure. These effects are not fully observable from proprioceptive signals alone. As shown in \cref{tab:variant_ablation}(a), replacing motion vectors with robot states (\textbf{M+M} $\rightarrow$ \textbf{S+S}) leads to a clear performance drop. This result indicates that motion vectors provide complementary dynamic information beyond robot states and play an important role in capturing interaction-driven visual dynamics.

\begin{table}[t]
\centering
\begin{tabular}{c|lcc}
\toprule
&\textbf{Ablation} & \textbf{Variant} & \textbf{SR} \\
\midrule
\multirow{3}{*}{(a)}&\multirow{3}{*}{\makecell[l]{Motion or State as\\(Hindsight+Foresight)}}
& S+M & 92.6 \\
&& S+S & 92.0 \\
&& M+M* & \textbf{94.4} \\
\midrule
\multirow{3}{*}{(b)}&\multirow{3}{*}{\makecell[l]{Causal or Bidirectional}}
& Causal-[M$|$A] & 87.4 \\
&& Bi-[A$|$M] & 94.0 \\
&& Bi-[M$|$A]* & \textbf{94.4} \\
\midrule
\multirow{2}{*}{(c)}&\multirow{2}{*}{Motion Representation} & Flow & 94.2 \\
&&MVs&\textbf{94.4}\\
\bottomrule
\end{tabular}
\caption{Ablation on different variants. M: Motion, S: State, A: Action.  * denotes the submission setting. (third view)}
\label{tab:variant_ablation}
\end{table}

\textbf{Bidirectional Interaction vs. Causal Separation.} To evaluate the benefit of joint foresight–action modeling, we compare our design with a decoupled variant that removes the bidirectional interaction between motion and action tokens. In this variant, motion and action tokens are processed using causal attention and decoded with separate prediction heads. This modification preserves parallel computation but prevents motion and action representations from interacting during inference. In contrast, our Joint Expert employs bidirectional attention to model motion and action synchronously, allowing the two representations to influence each other during the forward pass. As shown in \cref{tab:variant_ablation}(b), the decoupled causal variant achieves a success rate of 87.4\%, whereas our joint formulation improves performance to 94.4\%. This result suggests that enabling bidirectional interaction between foresight motion and action representations is important for effective decision making. We further investigate whether the ordering of motion and action tokens affects performance. Since bidirectional attention provides a global receptive field within each segment, the token order should not influence the learned representation. Consistent with this expectation, swapping the token order (M$|$A $\rightarrow$ A$|$M) results in less than 0.5\% performance change.

\textbf{Different Motion Representations.} We further study the impact of different motion representations by replacing motion vectors (MVs) with optical flow, as reported in \cref{tab:variant_ablation}(c). Optical flow provides dense motion estimation and serves as a commonly used alternative motion cue. Specifically, optical flow is estimated using RAFT~\cite{teed2020raft}. The results show that both representations achieve nearly identical success rates, indicating that the HiF architecture is robust to the specific choice of motion representation. This suggests that the performance gains primarily stem from the proposed modeling framework rather than a particular motion signal. However, the two representations differ significantly in computational cost. Computing optical flow introduces substantial preprocessing overhead. With a 4-frame history, flow-based inference requires 186.8 ms, whereas our MV-based approach takes only 121.6 ms. The efficiency gap becomes larger when the history length increases, resulting in up to a 78\% reduction in latency overhead with MVs for 8-frame histories. These results indicate that motion vectors provide a more efficient motion representation while maintaining comparable performance.

\begin{figure}[ht]
    \centering
    \includegraphics[width=1\linewidth]{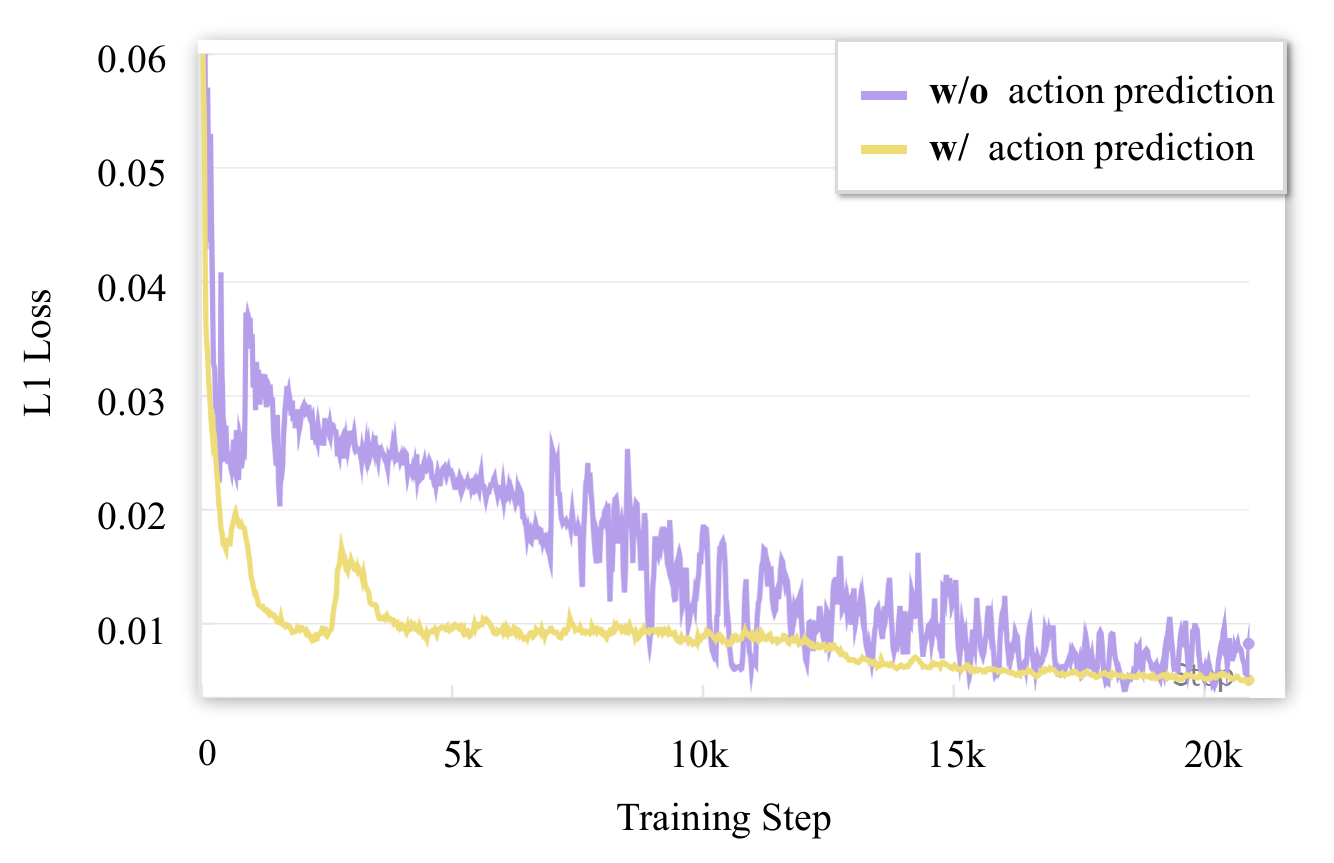}
    \caption{Convergence of the foresight-motion L1 loss during training. The curve ``w/o action prediction” corresponds to a variant where the action-prediction branch is removed and only the foresight-motion pathway is trained. The curve ``w/ action prediction” represents the full HiF-VLA architecture, where both streams in the joint expert (foresight motion and action) are retained.}
    \label{fig:wo_action}
\end{figure}

\subsection{Analysis of ``Think-while-Acting'' Paradigm}

\textbf{Foresight Motion Prediction.} To more comprehensively evaluate the interaction between foresight motion and action prediction, we additionally removed the action-prediction branch and trained the model using only the motion prediction pathway. We then compared the evolution of the motion L1 loss during training, as shown in \cref{fig:wo_action}. The results indicate that incorporating action prediction leads to faster convergence and a noticeably more stable training curve for motion prediction. This indicates that the action branch provides effective complementary information to the motion branch, enabling a synergistic interaction between the two. Such coupling supports the model’s ability to reason about future dynamics while simultaneously making action decisions—thereby achieving a genuine ``think-while-acting” capability.

\begin{figure*}
    \centering
    \includegraphics[width=1\linewidth]{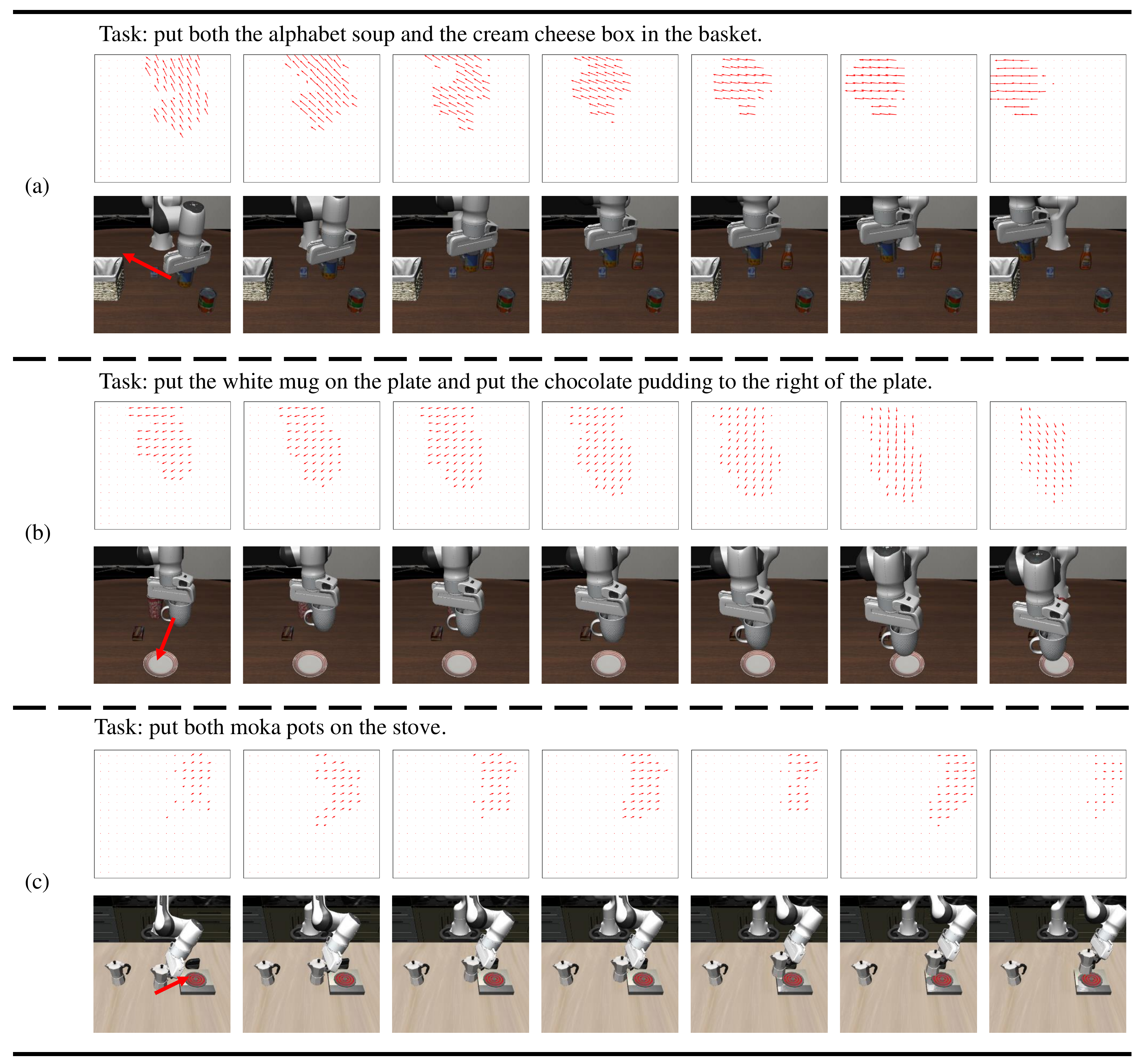}
    \caption{Example rollouts of three tasks in LIBERO-Long, illustrating the close alignment between the predicted foresight motion and the observed action execution over a short inference window.}
    \label{fig:motion_vis}
\end{figure*}

\textbf{Foresight and Action Prediction Visualization.} We present qualitative visualizations of the foresight motion predictions and action predictions across several tasks in the LIBERO-Long dataset, as shown in \cref{fig:motion_vis}. 
The visualizations demonstrate how our model's anticipated motion sequences closely align with its generated action plans. Across diverse long-horizon tasks, this close alignment ensures that the agent's actions are consistently grounded in a coherent forecast of future states.
This visually corroborates that HiF-VLA maintains temporal coherence and successfully executes the proposed ``think-while-acting'' paradigm by dynamically adapting its policy based on foresighted reasoning.

\section{Real-World Experiments}
\subsection{Real-World Experimental Setup}

We evaluate our method on a series of long-horizon real-world tasks using an AgileX Piper robot, which is equipped with a 6-DoF manipulator and a 1-DoF gripper. A single Intel RealSense D435 camera provides third-person observations, while an additional USB wrist-mounted camera provides egocentric input. Data are collected at 20 Hz. All models are trained under the standard HiF-VLA configuration and deployed on a single NVIDIA RTX 4090 GPU.

\subsection{Task Descriptions}

\textbf{Place blocks on the plates.} The robot must place the white block onto the white tray and the pink block onto the pink tray. This task primarily evaluates the model’s basic visual recognition ability and precise object placement. Its clear goal structure and low action complexity make it suitable for assessing whether the model can reliably establish accurate observation–action mappings in simple environments.

\textbf{Cover block and stack bowls.} The robot first covers a white block with a green bowl and then stacks a pink bowl on top of the green one. This task stresses the model’s ability in multi-step sequential reasoning, spatial relation understanding, and long-horizon dependency modeling. The explicit hierarchical dependency structure (cover $\xrightarrow{}$ stack) makes it an effective test for the model’s capacity to handle layered object interactions and long-term constraints.

\textbf{Press buttons in order.} The robot must press three colored buttons in a specified order. This task assesses the model’s ability to distinguish visually similar states, maintain correct action ordering, and perform time-sensitive decision making. Because pre-press and post-press observations can look visually similar, the task naturally introduces visual ambiguity and demands strong historical information integration and temporal reasoning. 

\subsection{Real-World Execution Visualization}

\begin{figure*}[htbp]
    \centering
    
    \begin{subfigure}[b]{0.99\textwidth}  %
        \centering        \includegraphics[width=\textwidth]{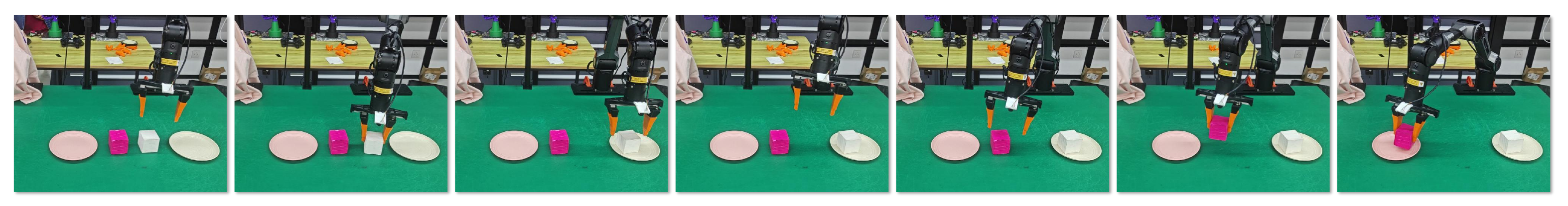}  %
        \caption{Place blocks on the plates.}
        \label{fig:a1}
    \end{subfigure}
    \begin{subfigure}[b]{0.99\textwidth}
        \centering        \includegraphics[width=\textwidth]{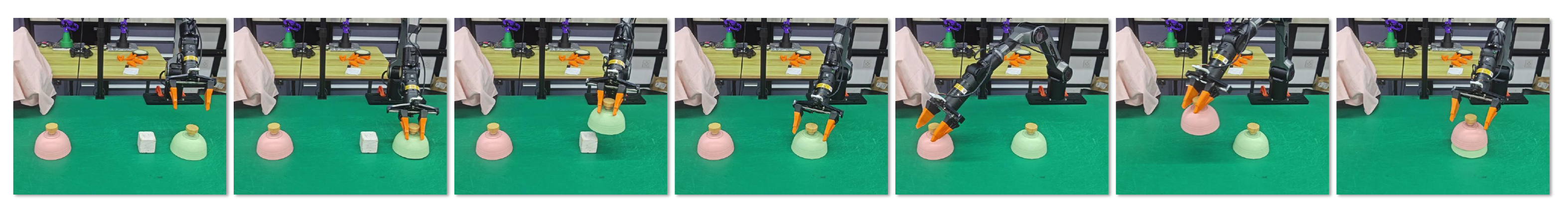}  %
        \caption{Cover block and stack bowls.}
        \label{fig:b1}
    \end{subfigure}
    \begin{subfigure}[b]{0.99\textwidth}
        \centering        \includegraphics[width=\textwidth]{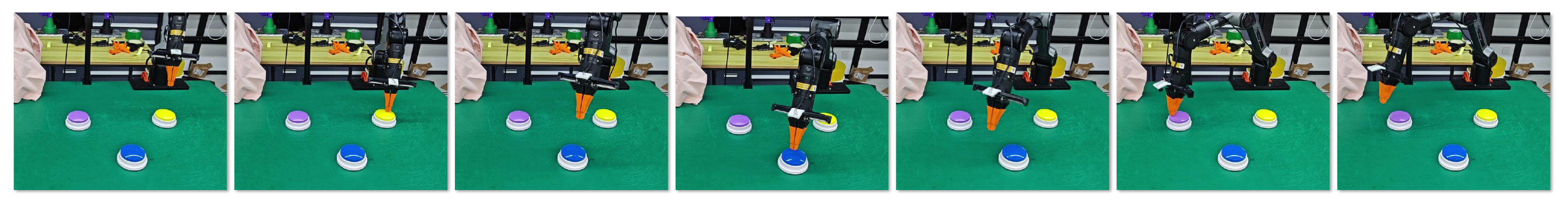}  %
        \caption{Press buttons in order.}
        \label{fig:c1}
    \end{subfigure}
     \vspace{-8pt}
    \caption{Example rollouts of real-world tasks.}
    \label{fig:real}
    \vspace{8pt}
\end{figure*}

\cref{fig:real} provides visualizations of HiF-VLA’s rollout across the real-world tasks. The figure presents the model’s generated actions at key time steps. As shown, HiF-VLA demonstrates stable behavior planning capabilities in real operation: in the block-grasping task, the robot consistently maintains reliable object recognition and performs precise grasp-and-place actions; in the multi-step bowl covering and stacking task, the system robustly executes cross-object sequential operations and preserves coherent action structure throughout stages with clear hierarchical dependencies; in the button-pressing task, the model correctly distinguishes between visually similar states and adheres to the required action order. These visualizations illustrate that HiF-VLA exhibits robust visual understanding, coherent action organization, and strong execution of long-horizon task structures in real scenarios, thereby validating its reliability in performing complex manipulation tasks.

\subsection{Failure Cases}
\begin{figure*}[htbp]
    \centering
    
    \begin{subfigure}[b]{0.99\textwidth}  %
        \centering        \includegraphics[width=\textwidth]{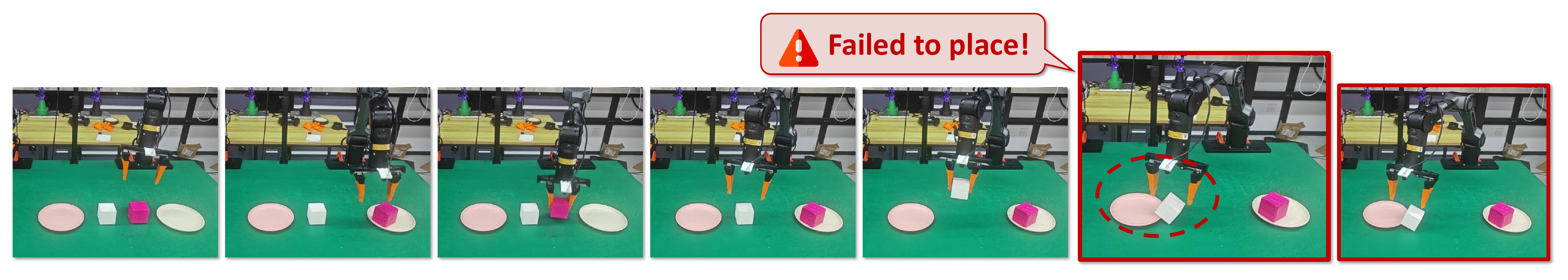}  %
        \caption{Place blocks on the plates.}
        \label{fig:a}
    \end{subfigure}
    \begin{subfigure}[b]{0.99\textwidth}
        \centering        \includegraphics[width=\textwidth]{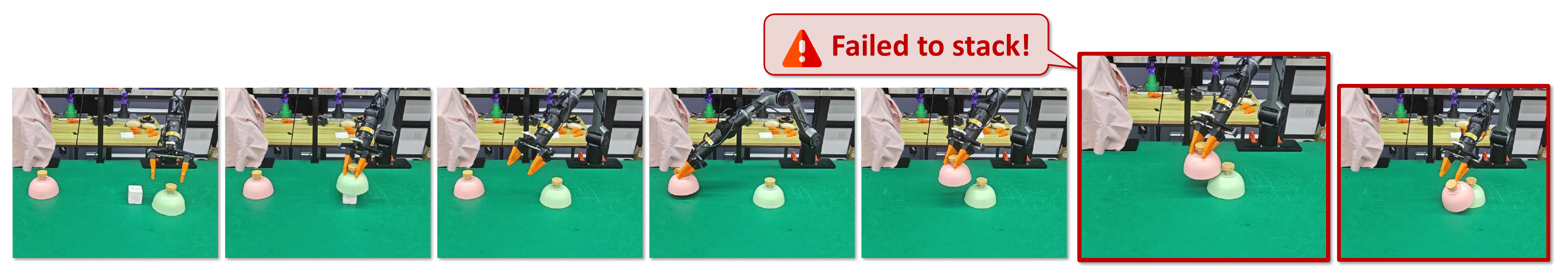}  %
        \caption{Cover block and stack bowls.}
        \label{fig:b}
    \end{subfigure}
    \begin{subfigure}[b]{0.99\textwidth}
        \centering        \includegraphics[width=\textwidth]{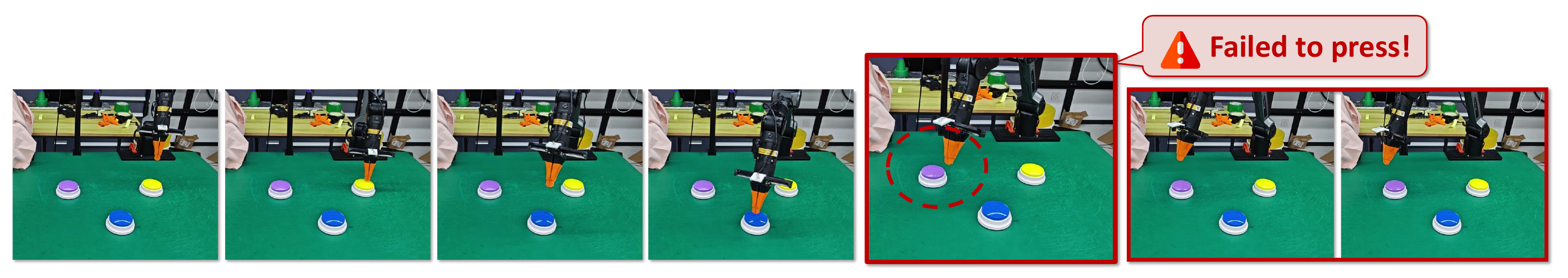}  %
        \caption{Press buttons in order.}
        \label{fig:c}
    \end{subfigure}
    \vspace{-8pt}
    \caption{Failure cases of real-world tasks.}
    \label{fig:fail_real}
\end{figure*}

Despite HiF-VLA outperforming in both simulation benchmarks and real-world experiments, several failure cases still occur, as shown in \cref{fig:fail_real}. In the first task, the model prematurely opened the gripper based on an incorrect spatial judgment, resulting in a placement failure. In the second task, the stacking failed because the robotic arm did not lift the bowl to an appropriate height. In the third task, an erroneous depth estimation caused the gripper to descend insufficiently, leading to an incomplete button press.

These failure modes highlight the critical role of spatial geometry and 3D perception. They suggest that future work may benefit from integrating richer 3D representations into our framework to further enhance robustness in real-world manipulation.

\end{document}